\documentclass[10pt,twocolumn,letterpaper]{article}

\usepackage{cvpr}              

\usepackage{graphicx}
\usepackage{amsmath}
\usepackage{amssymb}
\usepackage{booktabs}
\usepackage{rotating}
\usepackage[accsupp]{axessibility}  

\usepackage[pagebackref,breaklinks=true,colorlinks]{hyperref}

\usepackage{url}
\usepackage[capitalize]{cleveref}
\usepackage{xcolor}
\crefname{section}{Sec.}{Secs.}
\Crefname{section}{Section}{Sections}
\Crefname{table}{Table}{Tables}
\crefname{table}{Tab.}{Tabs.}

\def\cvprPaperID{*****} 
\def\confName{CVPR}
\def\confYear{2023}

\begin{document}

\title{
Towards Sim-to-Real Industrial Parts Classification with Synthetic Dataset
}

\author{Xiaomeng Zhu$^{1,3,}$\thanks{Corresponding author} , Talha Bilal$^{1}$, Pär Mårtensson$^{1}$, Lars Hanson$^{2}$, Mårten Björkman$^{3}$, Atsuto Maki$^{3}$\\
 $^{1}$Scania CV AB,
 $^{2}$University of Skövde,
 $^{3}$KTH Royal Institute of Technology\\
\small \texttt{\{xiaomeng.zhu, talha.bilal, par.martensson\}@scania.com,}\\ \small\texttt{\{xiazhu, celle, atsuto\}@kth.se, lars.hanson@his.se}
}
\maketitle


\begin{abstract}
This paper is about effectively utilizing synthetic data for training deep neural networks for industrial parts classification, in particular, by taking into account the domain gap against real-world images. To this end, we introduce a synthetic dataset that may serve as a preliminary testbed for the Sim-to-Real challenge; it contains 17 objects of six industrial use cases, including isolated and assembled parts. A few subsets of objects exhibit large similarities in shape and albedo for reflecting challenging cases of industrial parts. All the sample images come with and without random backgrounds and post-processing for evaluating the importance of domain randomization. We call it Synthetic Industrial Parts dataset (SIP-17). 
We study the usefulness of SIP-17 through benchmarking the performance of five state-of-the-art deep network models, supervised and self-supervised, trained only on the synthetic data while testing them on real data. By analyzing the results, we deduce some insights on the feasibility and challenges of using synthetic data for industrial parts classification and for further developing larger-scale synthetic datasets. Our dataset \footnote[2]{Dataset: \url{https://www.kaggle.com/datasets/mandymm/synthetic-industrial-parts-dataset-sip-17}} and code \footnote[3]{Code: \url{https://github.com/BilalTalha/SIP-17}} are publicly available.
\end{abstract}

\section{Introduction}
\label{sec:intro}

Efficient and reliable automatic parts classification is critical for various industrial operations and handling processes, such as sorted storing, part feeding, and quality inspection. With the increasing variability of products and required flexibility of processes and material flow, the importance of it has further escalated \cite{krueger2019deep}. Deep learning-based classification algorithms, with their robustness, can be a possible solution for industrial parts classification. However, training these algorithms generally requires a large amount of annotated data, which can be time-consuming and label-expensive to obtain in many real-world industry scenarios.

Synthetic data may present a viable solution to overcome this challenge. In the manufacturing industry, Computer-Aided Design (CAD) models are commonly used to create detailed virtual representations of physical objects for planning and simulating the manufacturing process \cite{tovey1989drawing}. Accordingly, synthetic data generated from CAD models can be useful to tackle the challenge of limited real-world data \cite{cohen2020cad,dekhtiar2018deep,wong2019synthetic}. However, a major issue in it is the domain gap between CAD data and real data, as they are derived from different distributions.

Numerous deep learning studies have focused on addressing the challenge of domain shift from simulated to real images, and a majority of them evaluate their models on the benchmark Sim-to-Real dataset, such as the Visual Domain Adaptation Dataset (VisDa) \cite{peng2018visda}. However, since the current benchmark datasets often consist of general objects such as animals, furniture, and street view, they may not adequately model the characteristics of industrial parts \cite{mayershofer2020loco}. In particular, the industrial environment often involves parts with subcategories differences or alignment variations that may not typically be captured by those datasets. As a result, the methods that perform well on the existing Sim-to-Real datasets may not generalize effectively to industrial scenarios.

Therefore, in this study, we introduce a Synthetic Industrial Parts dataset (SIP-17) which contains 17 objects representing six industrial use cases of parts sorting and quality inspection. The dataset comprises both isolated and assembled parts, some of which exhibit significant similarities or albedo, reflecting the challenges encountered in real-world industrial parts classification scenarios. As such, this dataset may serve as a preliminary testbed for Sim-to-Real industrial parts classification research. Testing new models on this dataset may also provide insights into the robustness of the model in solving various Sim-to-Real industrial parts classification use cases. 

The dataset focuses on the Sim-to-Real challenge, where only synthetic data is used for the training and validation (source) domains, while real images are used for the test (target) domain. Specifically, the dataset includes 66K labeled synthetic images for training and validation, and 566 unlabeled real images for testing. Unlike previous Sim-to-Real object identification datasets that are often benchmarked with domain adaptation models requiring real data for training \cite{peng2018visda,zhang2021survey}, we train our dataset only on synthetic data to enhance its practical value for industrial applications. By doing so, the manufacturers may bypass the need for manual data collection and annotation, and potentially develop parts classification models for quality inspection or parts sorting stations before the physical production of the parts.

Regarding the aforementioned domain gap, domain randomization can be a possible technique for addressing it. Tobin et al. \cite{tobin2017domain} introduced the concept of domain randomization which involves randomizing various aspects of the training data, such as camera positions, lighting conditions, object positions, and textures, to simulate a wide range of possible scenarios. The goal is to narrow the Sim-to-Real gap by generating synthetic data with sufficient variation allowing the model to perceive real-world data just as another variation\cite{tobin2017domain}. In this study, we generated the SIP-17 dataset following the domain randomization technique. To assess the impact of domain randomization, we generated the synthetic data with random backgrounds and random post-processing (Syn\_R) and without them (Syn\_O).

We evaluated state-of-the-art classification models on both Syn\_R and Syn\_O to establish benchmarks for our dataset. We selected a range of classification models with varying design principles, including Convolutional Neural Networks (CNNs), a Vision Transformer (VIT) \cite{dosovitskiy2020image}, and a self-supervised learning network.  The results demonstrated varying levels of performance while training on data from different use cases, providing insights into the feasibility and challenges of utilizing synthetic data for industrial parts classification. It may also indicate some direction for the development of a larger-scale synthetic dataset in the future. 

\section{Related Work}
\label{sec:relatedwork}

\subsection{Sim-to-Real Dataset}
Numerous datasets have been developed for Sim-to-Real tasks in the past. The Linemod \cite{hinterstoisser2013model} and Linemod-Occluded \cite{brachmann2014learning} datasets, for example, are widely used in 6D pose estimation in robotics. They include synthetic and real images of 15 general objects with varying textures, shapes, camera poses, lighting conditions, occlusions, and more. These datasets serve as the benchmarks for Sim-to-Real object localization and pose estimation tasks.

For Sim-to-Real classification tasks, the Visual Domain Adaptation Classification Dataset (VisDa-C) \cite{peng2018visda} is a benchmark dataset that comprises both synthetic and real images of 12 objects. The synthetic images are generated from 3D models rendered from various angles and lighting conditions, while the real images are sourced from the Microsoft COCO dataset \cite{lin2014microsoft} and the YouTube Bounding Box dataset \cite{real2017youtube}.

For Sim-to-Real segmentation tasks, multiple datasets have been introduced, particularly in the context of 2D and 3D multi-object tracking or autonomous guidance. These datasets often comprise synthetic images that are rendered from video games like GTA5 or different virtual urban environments, as well as real-world data obtained from a moving vehicle in urban settings or GPS, including RGB images, stereo images, and lidar data. Some examples of these datasets include the Domain Adaptation Segmentation Dataset (VisDa-S)\cite{peng2018visda}, the GTA5 dataset\cite{richter2016playing}, the Virtual KITTI \cite{gaidon2016virtual}, and KITTI \cite{geiger2013vision} datasets.

While many Sim-to-Real datasets feature general objects such as toys, animals, furniture, and street view, there are few options available for studying industrial objects. One of the datasets is the T-less dataset \cite{hodan2017t}, which includes 30 industrial objects with uniform textures and colors, such as bearings, U-brackets, metal boxes, and knives. Additionally, the Dataset of Industrial Metal Objects \cite{de2022dataset} offers real-world and synthetic multi-view RGB images of six objects, including cylinders, blocks, and shafts, placed on three different types of carriers: pallets, bins, and cardboard. However, these datasets are both designed for pose estimation, so they may not be well-suited for the challenge of cross-domain classification, as their test objects are limited to uniform or fixed textures and colors.

\subsection{Domain Randomization in Sim-to-Real}
Sadeghi and Levine \cite{sadeghi2016cad2rl} showed that quadcopters could be trained to fly indoors using only synthetic images, and Peng et al. \cite{peng2015learning} demonstrated the possibility of training object classifiers using 3D CAD models with random textures and backgrounds. Building on these ideas, Tobin et al. \cite{tobin2017domain} proposed the concept of domain randomization to address the reality gap by generating synthetic data with sufficient variations to enable the network to view real-world data as just another variation. Subsequent research \cite{tremblay2018training,prakash2019structured,yue2019domain} applied the domain randomization strategy to the GTA5 and Virtual KITTI datasets, training CNN-based object detection or segmentation models such as Faster-RCNN only on the synthetic data and achieving promising results while evaluating real-world data. 

For industrial parts identification, some works have utilized physics-based rendering and domain randomization to generate synthetic training data for various industrial parts, as demonstrated in two studies \cite{eversberg2021generating, horvath2022object}. The synthetic data is generated with randomized backgrounds, textures, post-processing, and other factors. Ablation studies are performed to analyze the impact of these factors on Sim-to-Real object detection tasks using object detection models such as Yolov4 \cite{bochkovskiy2020yolov4} and Faster R-CNN \cite{ren2015faster}. In these studies, Eversberg and Lambrecht \cite{eversberg2021generating} focus on identifying three types of turbine blades, while Horváth et al. \cite{horvath2022object} generated a dataset containing ten different objects, including bracket, pipe clamp, and handle. However, it is worth noting that the dataset only consists of isolated objects, which may not reflect the alignment differences that are frequently encountered in industrial assembly quality inspection use cases. In addition, the evaluation of the dataset using CNNs may not include a Sim-to-Real classification benchmark utilizing advanced models such as VIT.
\section{The SIP-17 Dataset}
\label{sec:SIP-17}

The SIP-17 dataset comprises 17 industrial objects that are representative of six use cases. The first four use cases consist of isolated parts that require classification. The last two involve assembled parts, where the objects consist of two or more parts that are assembled to each other, requiring inspection to ensure whether the parts are correctly aligned.  

\noindent\textbf{Use case 1: cabin assembly quality inspection}
Use Case 1 includes five objects: Airgun, Electricity12, Hammer, Hook, and Plug, which are assembled in the cabin of a truck. These objects from the same assembly station could share large similarities in albedo. By accurately classifying them, we can verify that the correct part has been assembled in the cabin.

\noindent\textbf{Use cases 2 to 4: logistic picking inspection}
Use case 2 comprises three objects: Fork1, Fork2, and Fork3; use case 3 includes four objects: CouplingHalf, Gear1, Gear2, and Pinion; and use case 4 consists of three objects: Cross, Pin1, and Pin2. These objects can be found in various logistic picking stations, and their classification is vital in ensuring that the operators have picked the correct parts for delivery. The parts in the same station are likely to belong to a closely related product family, which could share large similarities in shape and albedo.

\noindent\textbf{Use case 5: wheel assembly quality inspection}
Use case 5 involves inspecting whether a wheel has been correctly assembled with a screw. As the wheel can be assembled inside-out, leading to four possible categories during assembly: front side of the wheel with a screw (FwS), front side of the wheel without a screw (FwoS), back side of the wheel with a screw (BwS), and back side of the wheel without a screw (BwoS). 

\noindent\textbf{Use case 6: engine assembly quality inspection}
Use case 6 involves inspecting the Oring that is assembled on the Power Take Off. It includes three categories: correct assembly of the Orings (Oon), offside assembly of the Top Oring (Ooff), and missing Top Oring (noO). 

In summary, we selected four use cases with 15 isolated parts from the assembly and logistic stations. These parts varied in appearance, with some being similar while others differed.  Additionally, we included two use cases with two assembled parts. Each category of assembled parts shared numerous similarities as they contained the same objects but differed in alignment details. We chose these 17 objects from six use cases as they may represent the challenges encountered in real-world industrial parts classification scenarios. 

    \subsection{Dataset Acquisition}
To evaluate the effectiveness of domain randomization, we generated two synthetic datasets, one with random backgrounds and post-processing (Syn\_R) and one without those (Syn\_O). 

The Syn\_O dataset was created by rendering 3D CAD models from varying camera angles and under diverse lighting conditions. Each object was randomly rotated, scaled, and translated to generate variations. We used camera angles of 360 degrees and six distinct light directions for each model. To ensure that the entire object was captured, the camera was automatically positioned, and random light intensities were used during rendering. We followed the rendering parameters described in work \cite{volkswagen2022}.

The Syn\_R dataset was generated using a similar process as the Syn\_O dataset, with the additional step of introducing random colors to the lighting. Besides, we incorporated randomly selected backgrounds from the Unsplash dataset \cite{Unsplash}, along with various post-processing techniques such as random color tints, blurs, and noise \cite{volkswagen2022}. 

During the generation of the dataset, the randomization process was only applied to the virtual camera and environment. As for the CAD models, we applied a single-color texture that approximated the real objects to give them a reasonably realistic appearance. Nevertheless, to maintain the Sim-to-Real domain gap, other parameters affecting surface appearance, such as metallic, specular, and roughness, were held constant across all CAD models.

\begin{figure*}
  \centering
  \includegraphics[width=1\textwidth,trim=195pt 55pt 230pt 55pt,clip]{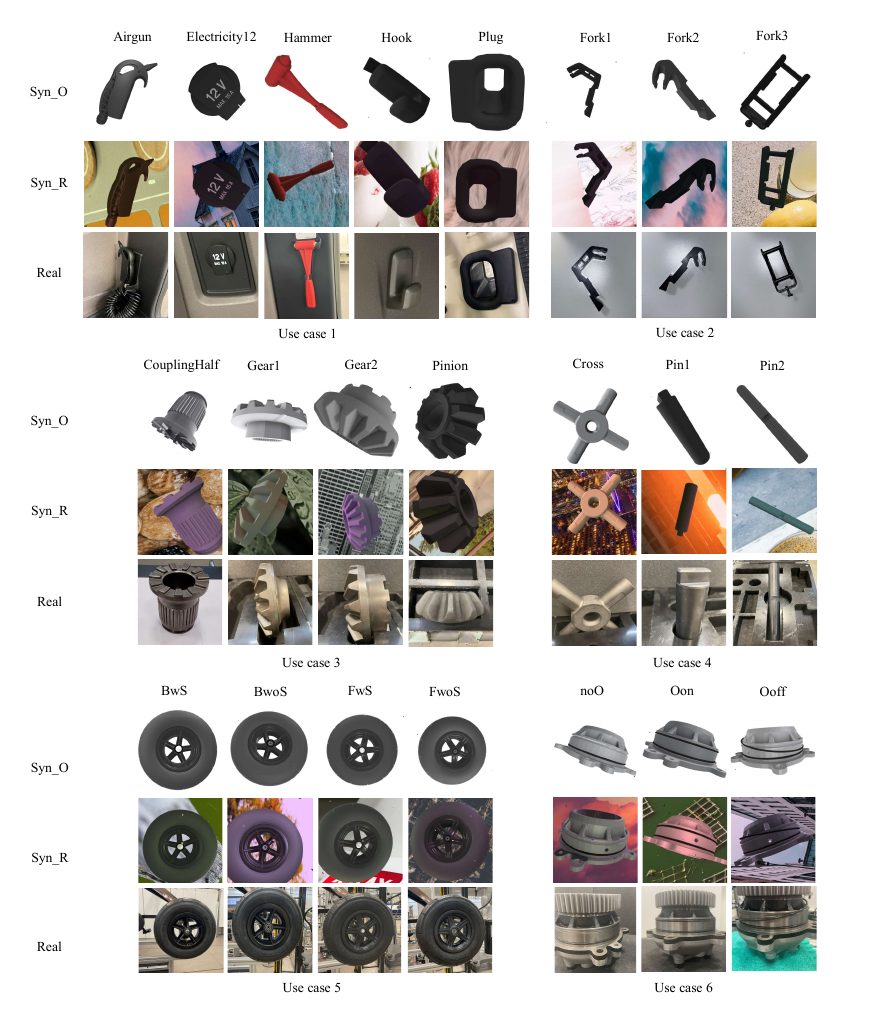}
   \caption{Sample images from the SIP-17 dataset, showcasing three categories: Syn\_O, synthetic images without random backgrounds and post-processing; Syn\_R, synthetic images with random backgrounds and post-processing; and Real, images captured from cameras in real industrial scenarios. Use cases 1-4 require the classification of isolated industrial parts, while use cases 5 and 6 require the classification of assembled parts. }
   \label{fig:dataset}
\end{figure*}

In total, we generated 33K images for both Syn\_R and Syn\_O datasets.  For each category in each use case, we generated 1200 synthetic images for training and 300 synthetic images for validation. For testing, we captured 566 real images from various industrial scenarios. The number of images per category is outlined in \cref{Table1}. Some samples of Syn\_R, Syn\_O, and real images for each category are present in \cref{fig:dataset}.

\begin{table}
  \centering
  \fontsize{14}{16}\selectfont
  \resizebox{\linewidth}{!}{%
  \begin{tabular}{@{}ccp{2cm}p{2cm}p{1.8cm}@{}}
    \toprule
    Use cases & Categories & Train (Syn\_R/ Syn\_O) & Valid (Syn\_R/ Syn\_O)  & Test (real images)\\
    \midrule
     & Airgun & 1200 &300 & 39 \\
     & Electricity12 & 1200 &300 & 44 \\
    Use case 1& Hammer & 1200 &300 & 34 \\
     & Hook & 1200 &300 & 40 \\
     & Plug & 1200 &300 & 53 \\ 
     \midrule
    & Fork1 & 1200 &300 & 32 \\
      Use case 2 & Fork2 & 1200 &300 & 30 \\
   & Fork3 & 1200 &300 & 30 \\
   \midrule
     & CouplingHalf & 1200 &300 & 33 \\
   Use case 3  & Gear1 & 1200 &300 & 34 \\ 
    & Gear2 & 1200 &300 & 38 \\
   & Pinion & 1200 &300 & 44 \\
   \midrule
     & Cross & 1200 &300 & 40 \\
     Use case 4 & Pin1 & 1200 &300 & 39 \\ 
     & Pin2 & 1200 &300 & 36 \\ 
     \midrule
     & Back\_wheel\_with\_screw (BwS) & 1200 &300 & 32 \\
   Use case 5 & Back\_wheel\_without\_screw (BwoS) & 1200 &300 & 32 \\ 
    & Front\_wheel\_with\_screw (FwS) & 1200 &300 & 32 \\
   & Front\_wheel\_without\_screw (FwoS) & 1200 &300 & 32 \\
   \midrule
    & Orings\_on (Oon) & 1200 &300 & 42 \\
     Use case 6 & TopOring\_off (Ooff) & 1200 &300 & 42 \\ 
     & no\_TopOring (noO) & 1200 &300 & 42 \\ 
    \bottomrule
  \end{tabular}
  }
  \caption{Number of images per category in the SIP-17 dataset.}
  \label{Table1}
\end{table}

\section{Evaluation}
\label{Evaluation}

\subsection{Experimental Setup}
Our evaluation of the SIP-17 dataset as a benchmark involves testing five classification models that are currently considered state-of-the-art. These models, which represent different design principles, are widely known for their effectiveness and relatively simple implementation. The five models under evaluation are:

\noindent\textbf{ResNet \cite{he2016deep}:} ResNet is a deep CNN architecture that introduced the concept of residual connections. Its simplicity and effectiveness have made it a popular baseline for image classification and benchmarking new methods. It is also frequently used as a classifier for various Sim-to-Real object identification tasks. In our experiments, we employed ResNet with 152 layers (ResNet152). 

\noindent\textbf{EfficientNet \cite{tan2019efficientnet}:} EfficientNet is a family of CNN architectures that achieve state-of-the-art performance while being computationally efficient. In our experiments, we utilized the EfficientNet B7 model.

\noindent\textbf{ConvNext \cite{liu2022convnet}:} ConvNext is a recent CNN architecture that introduced a split-attention mechanism to enhance the ability of the network to aggregate features. It "modernized" a standard ResNet toward the design of a Vision Transformer and achieved state-of-the-art classification results in CNNs on several benchmarks. In our experiments, we employed the ConvNext base model.

\noindent\textbf{Vision Transformer (VIT) \cite{dosovitskiy2020image}:} VIT is based on the transformer architecture used in natural language processing. Unlike traditional CNNs, it employs a self-attention mechanism that processes image patches directly, effectively capturing global dependencies and relationships between different parts of the input image. It has achieved state-of-the-art performance on several image recognition benchmarks. In our experiments, we utilized the VIT model with a base configuration and a patch size of 16 (vit\_b\_16). 

\noindent\textbf{DINO \cite{caron2021emerging}:} DINO is a self-supervised contrastive learning approach that improves feature representation for image classification tasks. It utilizes a teacher network to generate representations of an image and trains a student network to predict similarities between pairs of images in order to learn more meaningful and transferable features. Self-supervised learning methods have shown some effectiveness in Sim-to-Real tasks by learning features that are more transferable across domains. For example, Tian et al. \cite{tian2020contrastive} have proposed a self-supervised approach using contrastive learning to learn domain-independent features. In light of this, we aim to evaluate a self-supervised contrastive learning model DINO on our dataset. In our experiments, we employed the DINO model with VIT as its backbone and utilized a base configuration with a patch size of 16 (Dino\_vitbase16).

We chose the models with a comparable amount of parameters. For the self-supervised learning model DINO, we used the VIT pre-trained on ImageNet as its backbone and trained the linear classifier on our dataset for 25 epochs. For supervised learning models, we trained the models pre-trained on ImageNet for 25 epochs. To thoroughly evaluate the models performance, we conducted two types of experiments: (1) training the models with the 15 isolated parts and (2) training the models with objects per use case. All the models were trained with Syn\_R and Syn\_O datasets and tested on real images.

\subsection{Experimental Results and Discussion}
\begin{figure*}
  \centering
  \begin{subfigure}{0.36\linewidth}
  \centering
  \includegraphics[width=\textwidth]{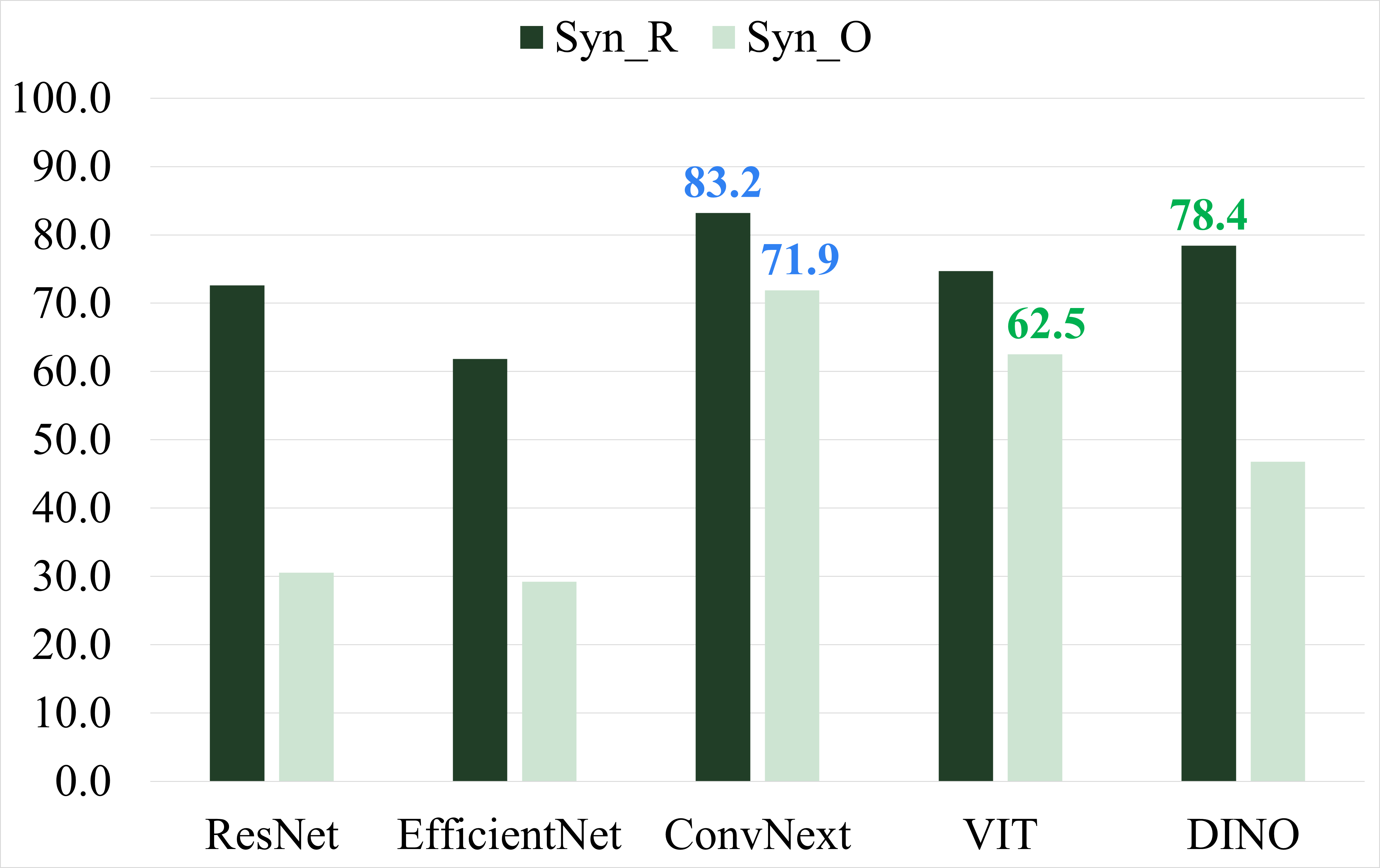}
    \caption{Accuracy (\%) of all isolated parts. Blue and green colors indicate the best and second-best models in terms of accuracy while training with Syn\_R or Syn\_O.}
    \label{fig:short-a}
  \end{subfigure}
  \hfill
  \begin{subfigure}{0,63\linewidth}
  \centering
  \includegraphics[width=\textwidth]{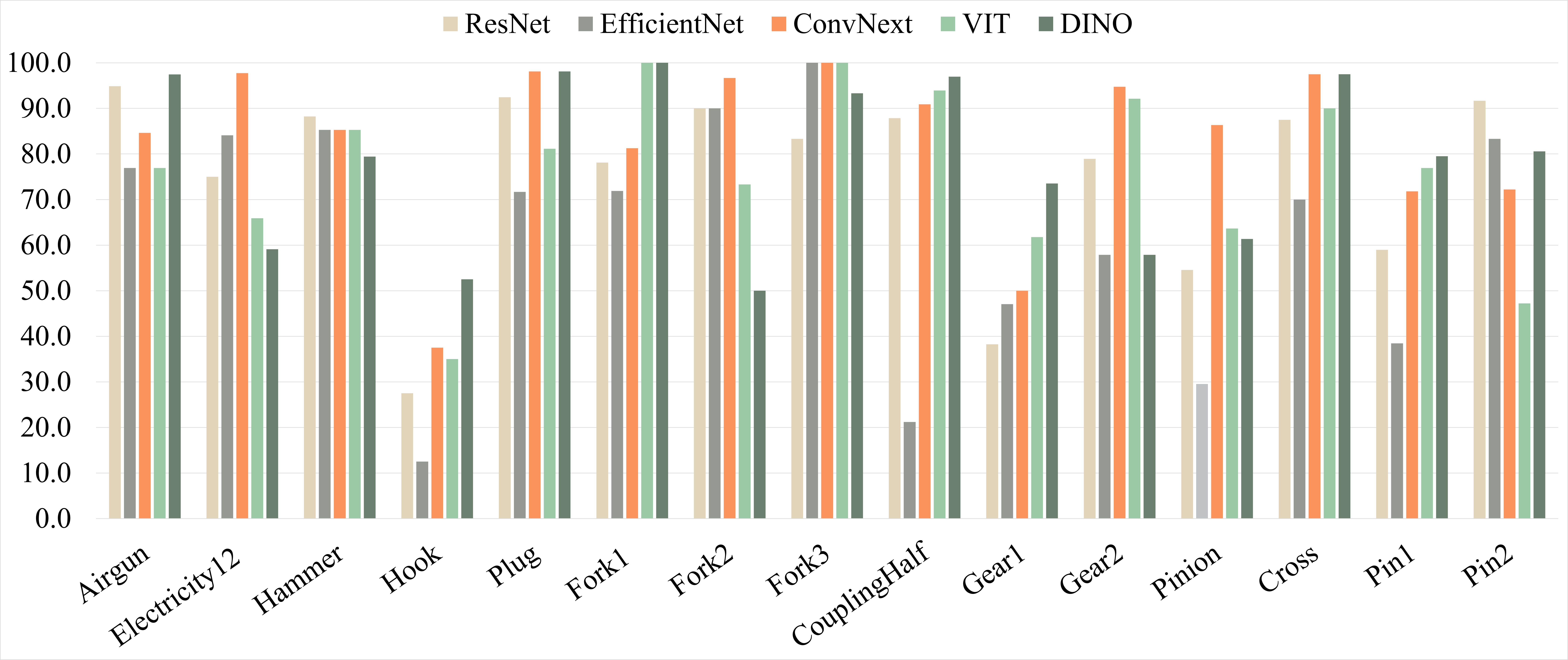}
    \caption{Class-wise accuracy (\%) of all isolated parts while trained with Syn\_R.}
    \label{fig:short-b}
  \end{subfigure}
  \caption{Results of all isolated parts.}
  \label{fig:isolated}
\end{figure*}

\begin{figure*}
  \centering
  \begin{subfigure}{0.494\linewidth}
  \centering
  \includegraphics[width=\textwidth]{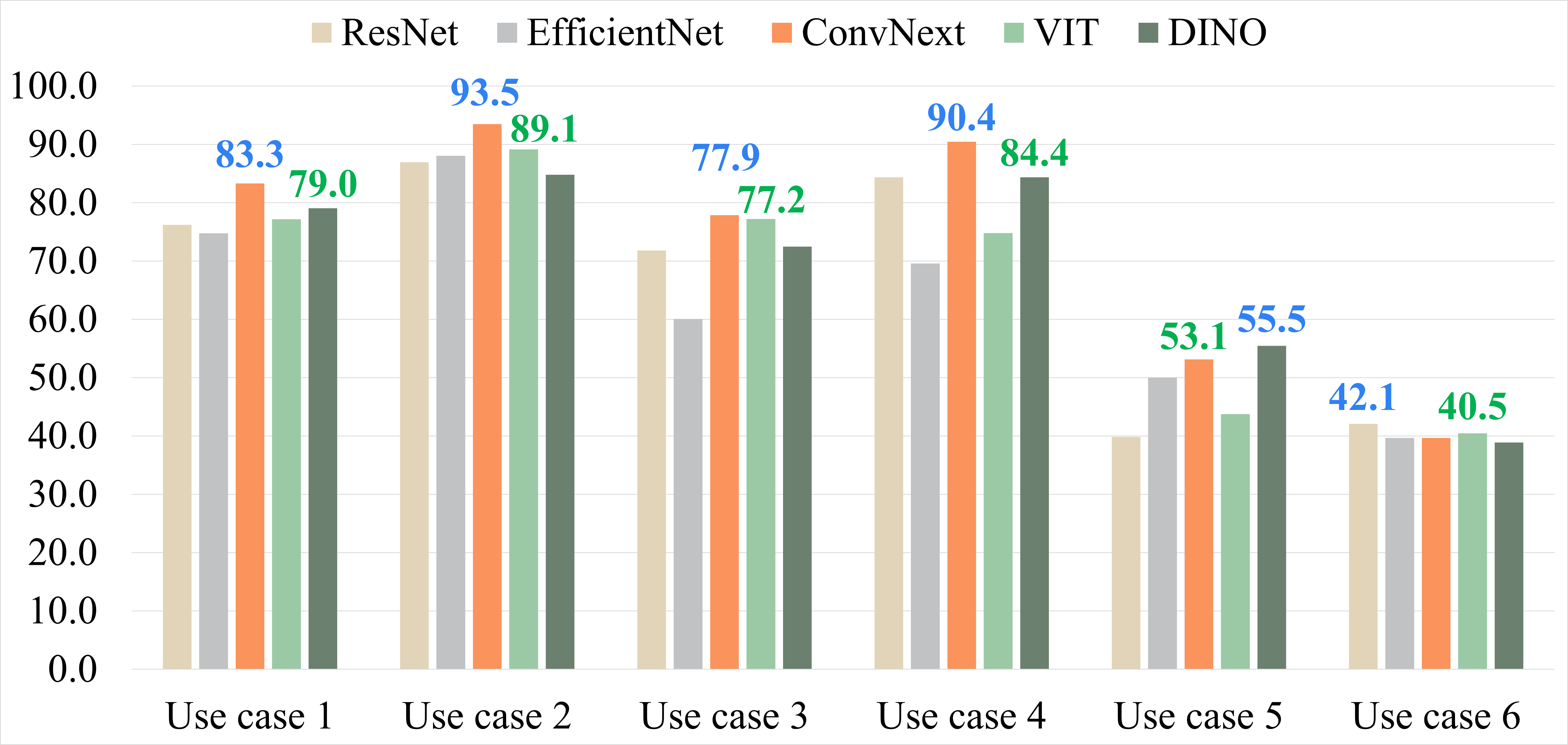}
    \caption{Accuracy (\%) of all use cases trained with Syn\_R. Blue and green colors indicate the best and second-best models in terms of accuracy.}
    \label{fig:short-a}
  \end{subfigure}
  \hfill
  \begin{subfigure}{0.494\linewidth}
  \centering
  \includegraphics[width=\textwidth]{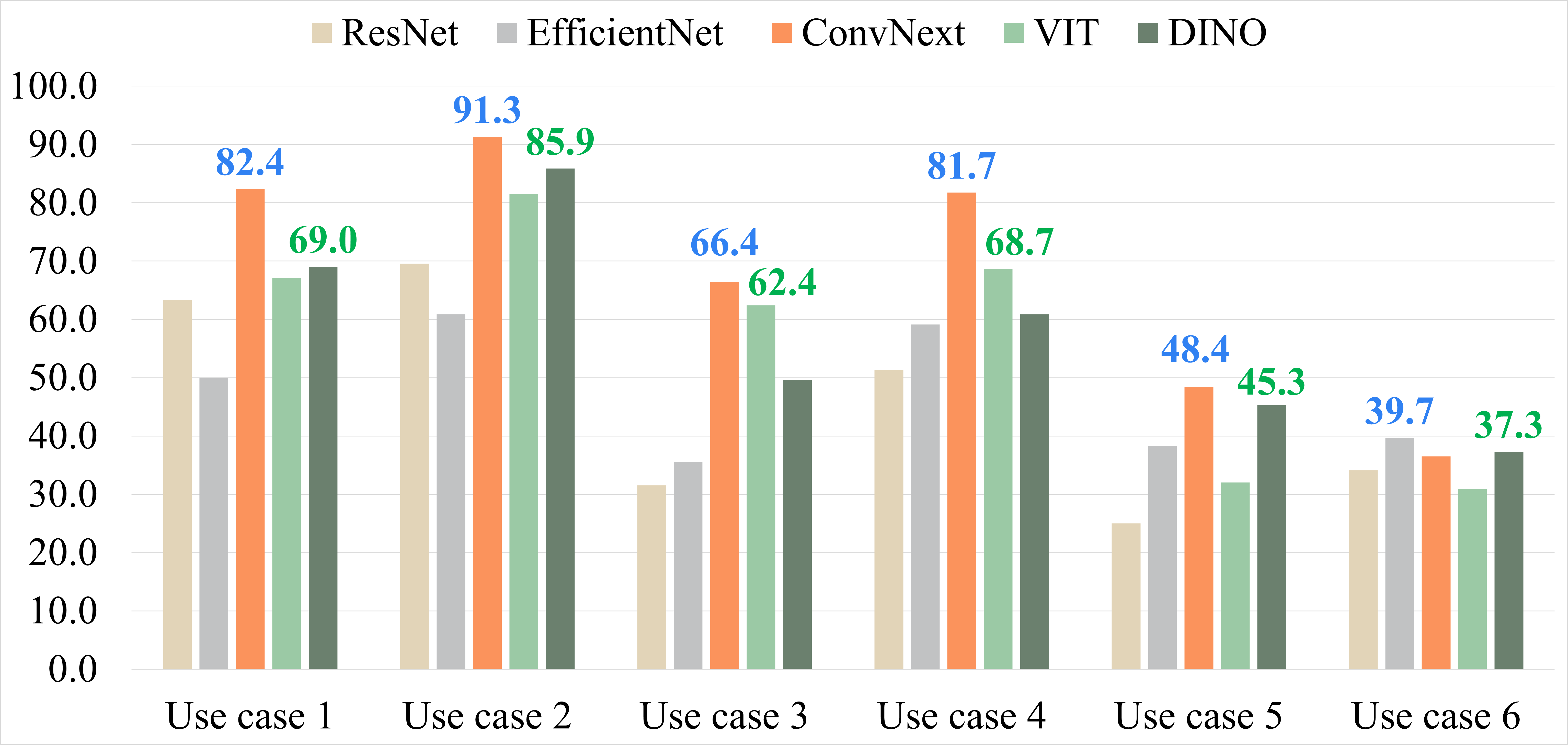}
    \caption{Accuracy (\%) of all use cases trained with Syn\_O. Blue and green colors indicate the best and second-best models in terms of accuracy.}
    \label{fig:short-b}
  \end{subfigure}
    \begin{subfigure}{1\linewidth}
  \centering
  \includegraphics[width=\textwidth]{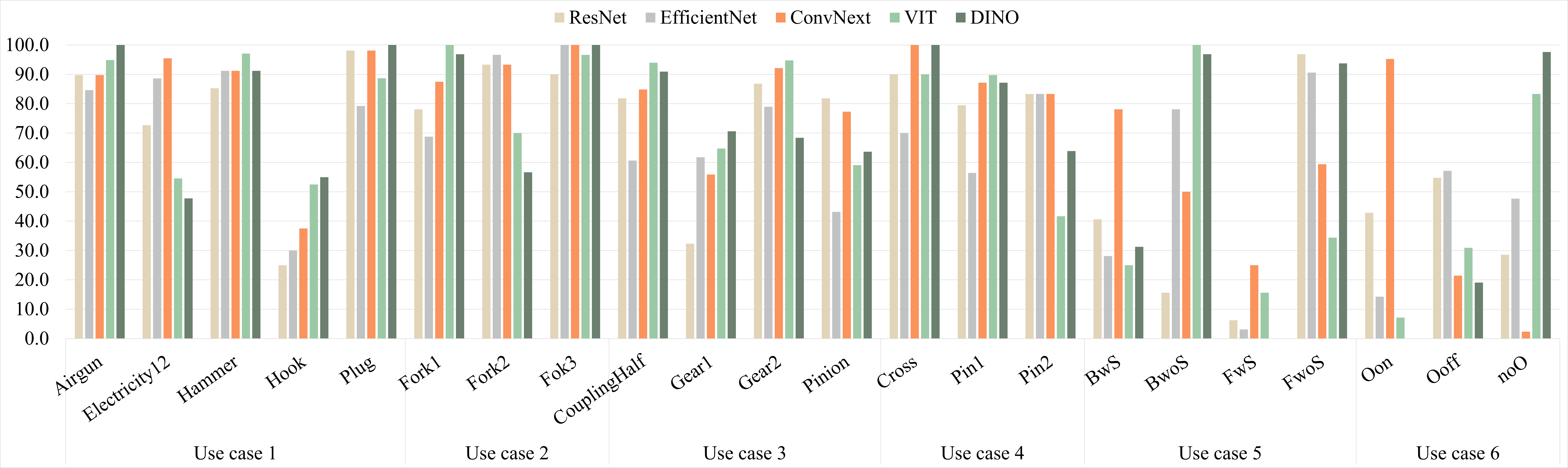}
    \caption{Class-wise accuracy (\%) of all use cases trained with Syn\_R.}
    \label{fig:short-c}
  \end{subfigure}
  \caption{Results of all use cases.}
  \label{fig:all}
\end{figure*}

 All the experiments have been repeated three times to obtain an average top-1 classification accuracy. The results of training on 15 isolated parts are presented in \cref{fig:isolated}, while the results of training on each individual use case are summarized in \cref{fig:all}. To highlight the best and second-best models in terms of total accuracy trained with Syn\_R and Syn\_O, we use blue and green colors, respectively, in \cref{fig:isolated} (a) as well as \cref{fig:all} (a) and (b). 

\begin{figure*}
  \centering
  \begin{subfigure}{0.32\linewidth}
  \centering
  \includegraphics[width=1\textwidth]{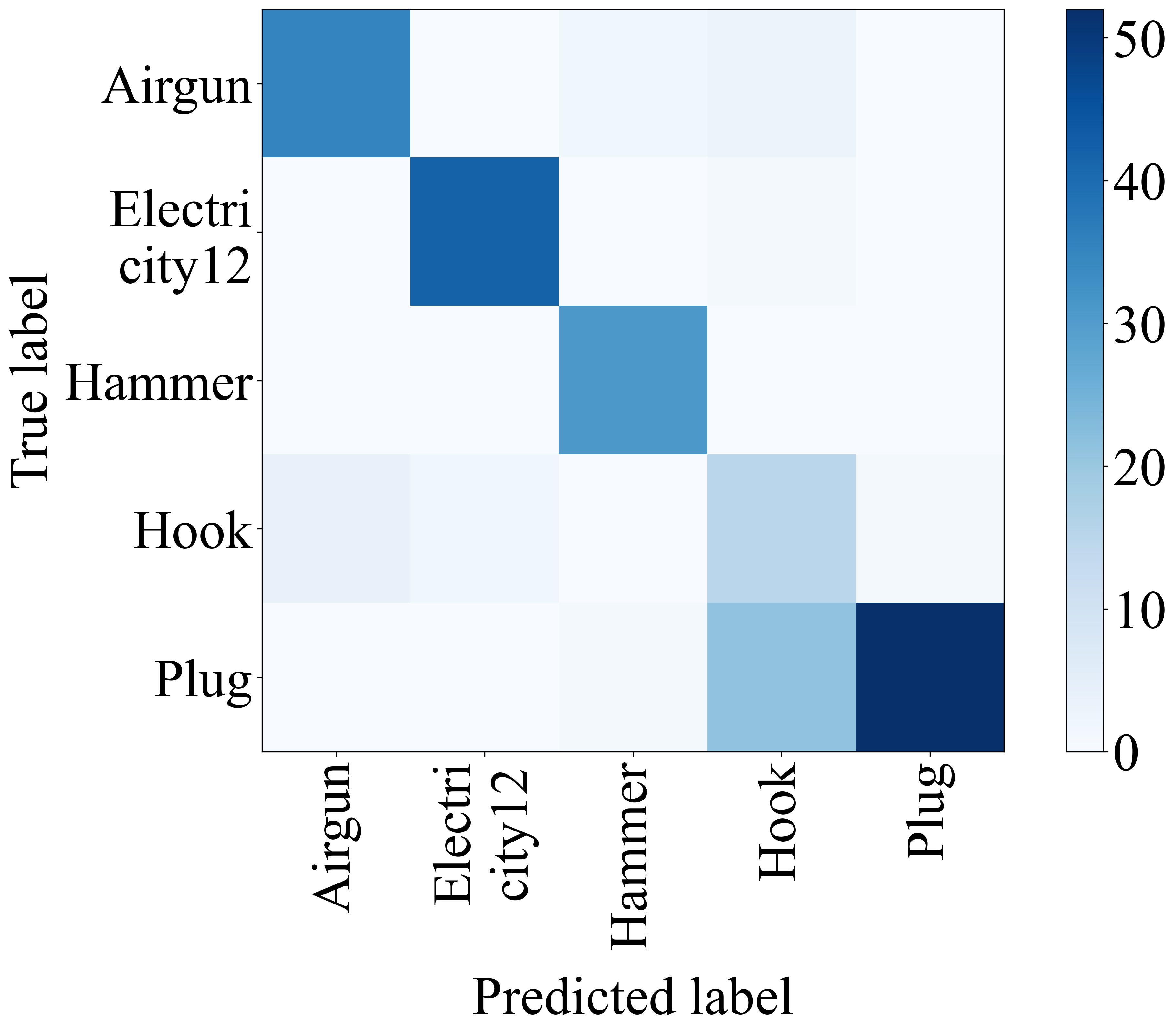}
    \caption{Confusion matrix on the use case 1. }
    \label{fig:short-a}
  \end{subfigure}
  \hfill
  \begin{subfigure}{0.32\linewidth}
  \centering
  \includegraphics[width=1\textwidth]{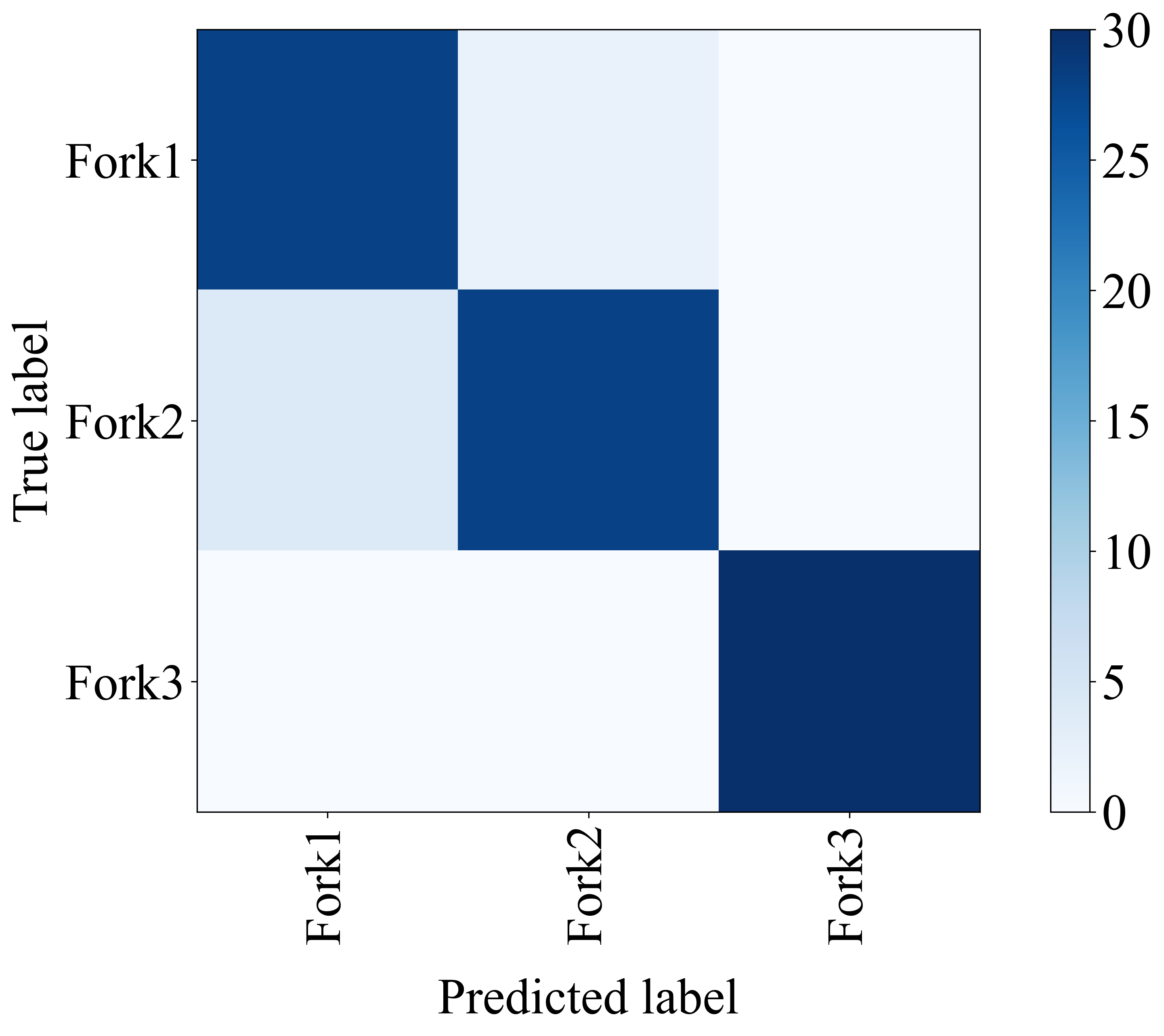}
    \caption{Confusion matrix on the use case 2. }
    \label{fig:short-b}
  \end{subfigure}
  \hfill
  \begin{subfigure}{0.32\linewidth}
  \centering
  \includegraphics[width=1\textwidth]{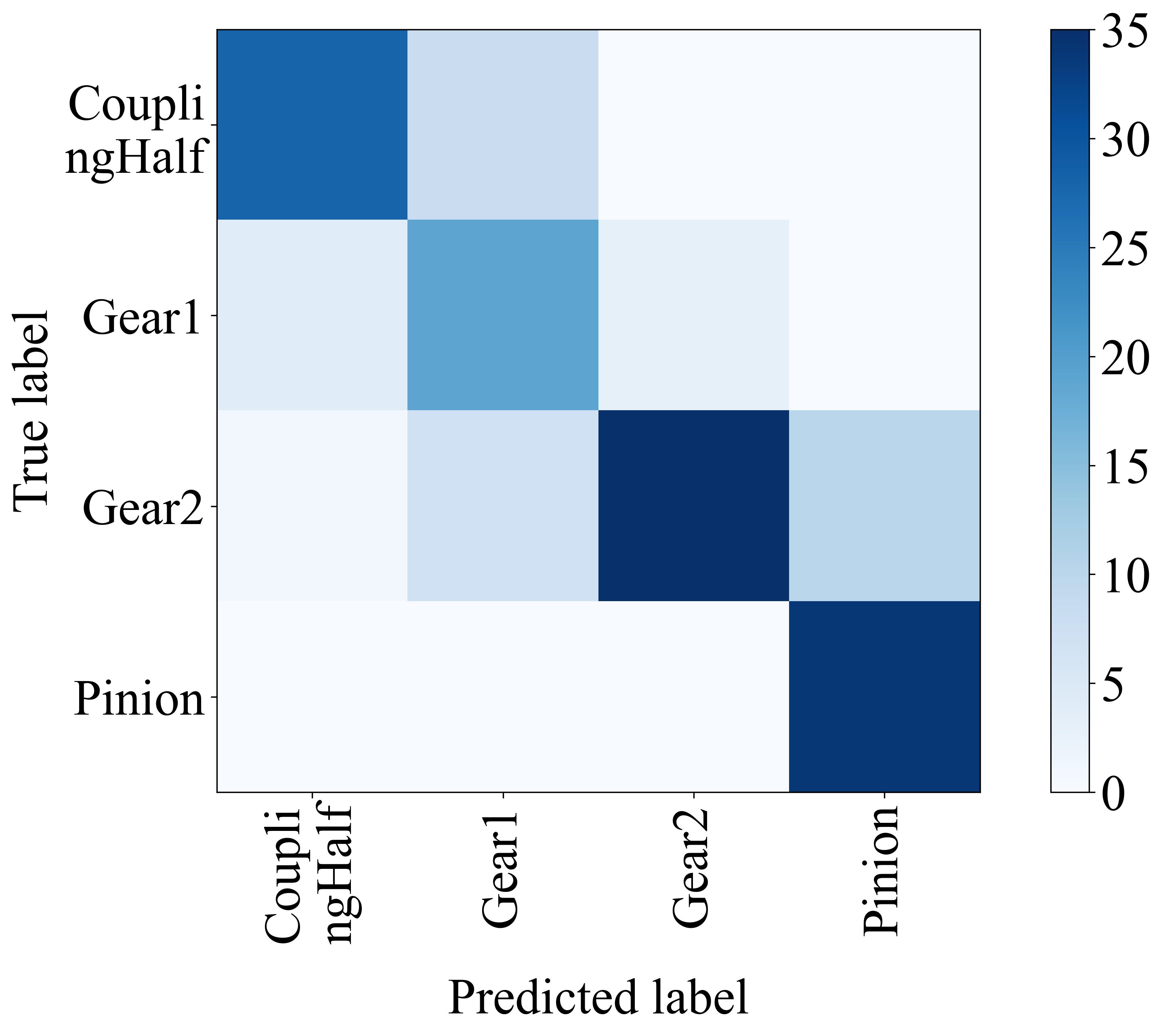}
    \caption{Confusion matrix on the use case 3. }
    \label{fig:short-c}
  \end{subfigure}
  \hfill
  \begin{subfigure}{0.32\linewidth}
  \centering
  \includegraphics[width=1\textwidth]{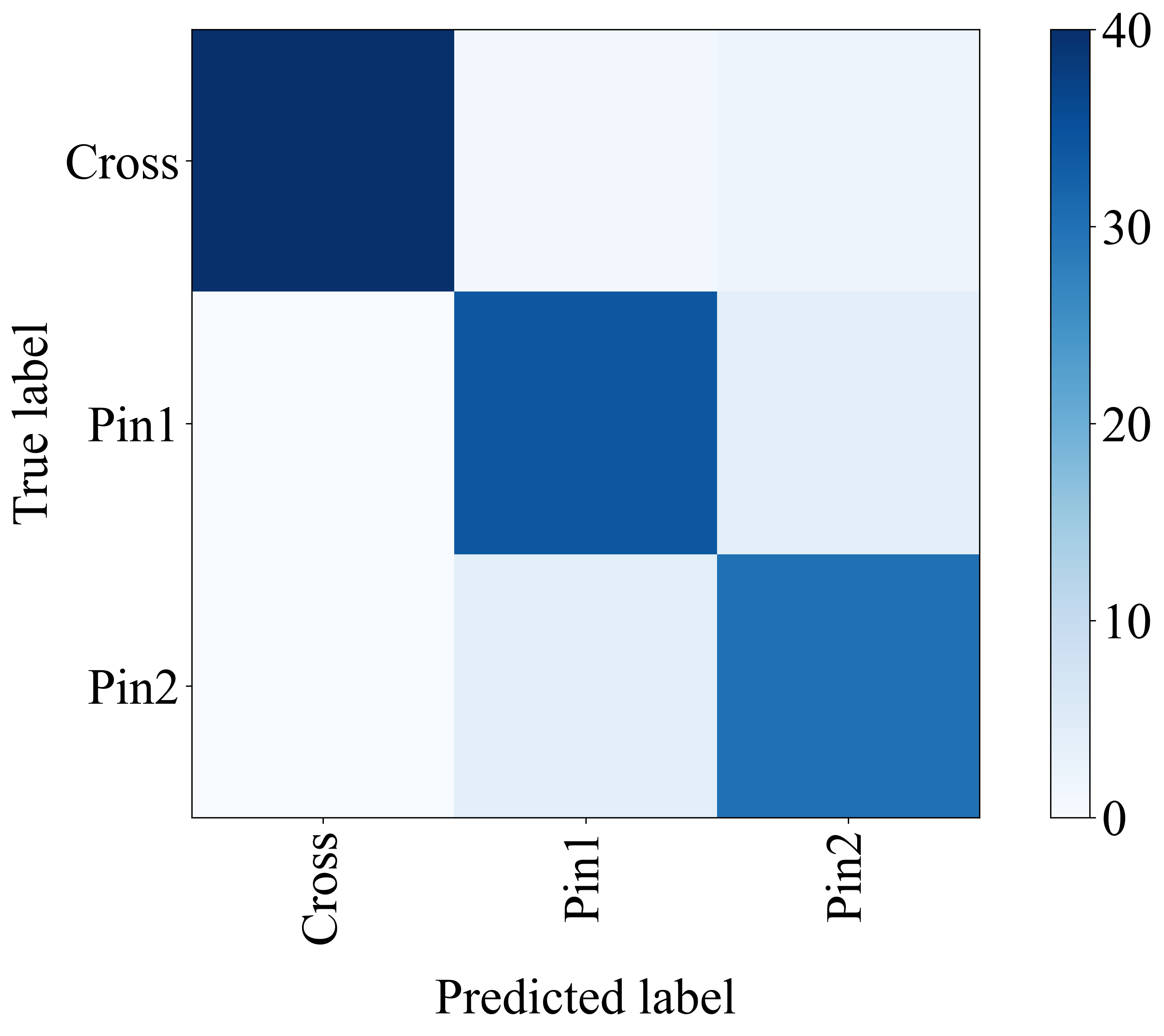}
    \caption{Confusion matrix on the use case 4. }
    \label{fig:short-d}
  \end{subfigure}
  \hfill
  \begin{subfigure}{0.32\linewidth}
  \centering
  \hspace{-0.65cm}
  \includegraphics[width=1\textwidth]{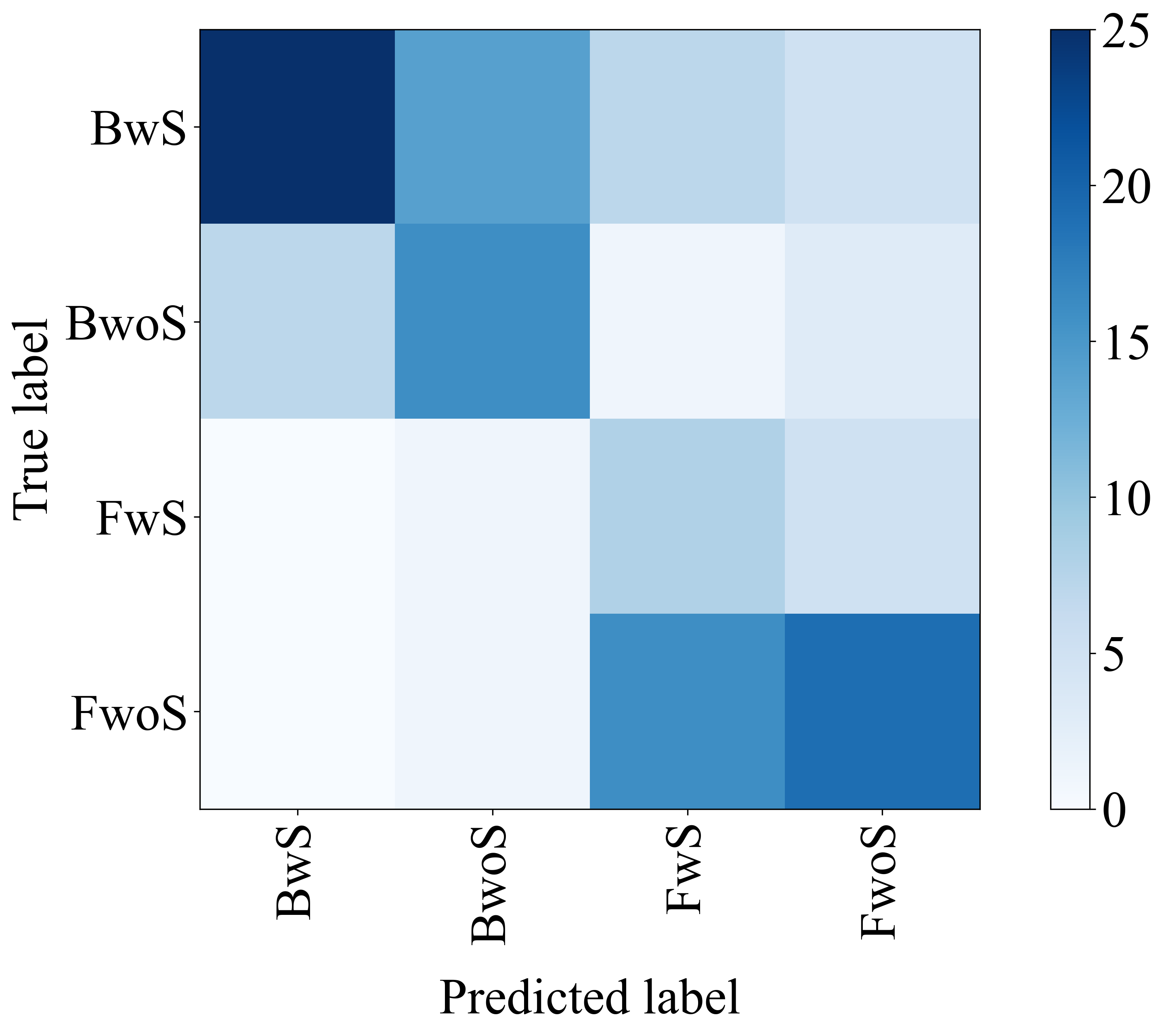}
    \caption{Confusion matrix on the use case 5. }
    \label{fig:short-e}
  \end{subfigure}
  \begin{subfigure}{0.32\linewidth}
  \centering
  \includegraphics[width=1\textwidth]{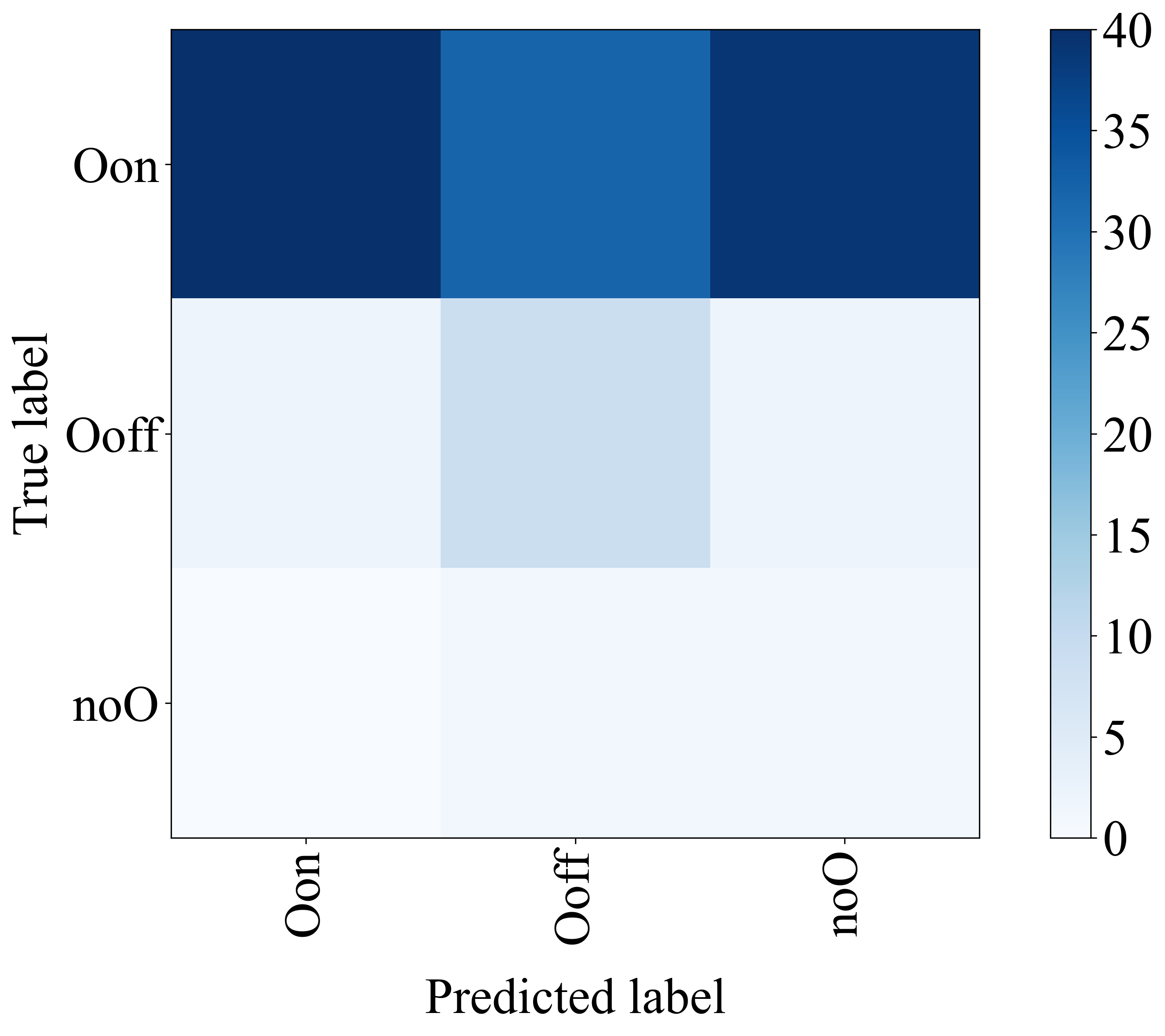}
    \caption{Confusion matrix on the use case 6. }
    \label{fig:short-f}
  \end{subfigure}
  \caption{Confusion matrices on different use cases with the ConvNext model.}
  \label{fig:Confusion_matrix}
\end{figure*}

\noindent\textbf{Domain randomization comparison:}
By comparing \cref{fig:isolated} (a), \cref{fig:all} (a) and (b), it is evident that the models trained on Syn\_R outperformed those trained on Syn\_O when trained on both 15 isolated parts and individual use cases. These findings suggest the significance of domain randomization in Sim-to-Real tasks, highlighting the benefits of training models with diverse synthetic data to improve their resilience to real-world variations.
As models trained on Syn\_R achieved better overall performance, our analysis will mainly focus on the results obtained from this dataset. We present the class-wise performance of the models trained with Syn\_R in \cref{fig:isolated} (b) and \cref{fig:all} (c).

\noindent\textbf{Model performance comparison:}     
As shown in \cref{fig:isolated}, training on 15 isolated parts with Syn\_R yielded the highest total accuracy of 83.2\% for the ConvNext model, followed by DINO (78.4\%) and VIT (74.7\%). These results are consistent with those in \cref{fig:all}, where ConvNext demonstrated the best performance in use cases 1 to 4 and the second-best performance in use case 5, followed by DINO and VIT. Notably, the ConvNext model exhibited the smallest performance difference between training on Syn\_O and Syn\_R, indicating its potential robustness for cross-domain classification, possibly due to its split-attention mechanism. On the other hand, the self-supervised learning model DINO achieved the second-best average performance, which may suggest the effectiveness of contrastive learning strategies for cross-domain classification tasks. 

The results also suggest that DINO outperformed ConvNext in some categories, such as Hook and Gear1. Therefore, combing the strength of ConvNext and DINO may potentially lead to further improvement in model performance. For instance, we could use a contrastive learning strategy on ConvNext or add a supervised loss to DINO to create supervised contrastive learning models with attention mechanisms, potentially enhancing their ability to capture subcategory details in cross-domain classification. 

\noindent\textbf{High-performing categories:}    The highest accuracy achieved in our experiments was by the ConvNext model in use cases 2 and 3 while trained with Syn\_R, with accuracies of 93.5\% and 90.4\%, respectively. These findings indicate the potential of utilizing synthetic data in parts classification and suggest that training models only on synthetic images with domain randomization may yield promising results.

\noindent\textbf{Low-performing categories}     We carried out further analysis to identify which categories may have contributed to the low performance in other use cases.
When analyzing the results of isolated parts classification from use cases 1 to 4, a comparison between \cref{fig:isolated} (b) and \cref{fig:all} (c) reveals that the class-wise results obtained from training on individual use cases exhibit a similar trend to those obtained from training on all 15 parts. These findings suggest that, in industrial parts classification, 
models are likely to confuse objects that share the same manufacturing process and stations.
In addition, both \cref{fig:isolated} (b) and  \cref{fig:all} (c) reveal that the categories Hook and Gear1 received the lowest accuracy across most models in isolated parts classifications. 

As for the results of assembled parts classification from use cases 5 and 6, \cref{fig:all} (a) indicates that they performed considerably worse than the isolated part classification. All models received an accuracy of around 50\% or lower in use cases 5 and approximately 40\% in use cases 6.

To further explore the issue, confusion matrices were generated for all use cases of the best model, ConvNext, as shown in \cref{fig:Confusion_matrix}. Upon analyzing these matrices, it became apparent that certain subcategories exhibited high confusion rates. Specifically, in the isolated parts classification, Hook was frequently confused with Plug, while Gear1 was often mistaken for CouplingHalf or Gear2. In the assembled parts classification of use case 5, Back\_wheel\_with\_srew (BwS) exhibits high confusion with Back\_wheel\_without\_srew (BwoS), and Front\_wheel\_with\_screw (FwS) was frequently misclassified as Front\_wheel\_without\_screw (FwoS). Moreover, in use case 6, all categories were confused and misclassified as Orings\_on (Oon). These low-performance results indicate that the models failed to capture the semantic representation that distinguishes these categories.

To summarize the issue, we divided these categories into two groups. The first group comprises objects with similar albedos and simple shapes, such as Hooks and Plugs. As depicted in \cref{fig:dataset} use case 1, Hooks are commonly assembled on a black surface in real images. Since the color of the Hook is also black, and it has a small size and simple shape, the model may focus on the albedo of the entire image rather than the specific features of the Hook, resulting in confusion with other objects that share similar albedo characteristics with Hook, such as Plugs.

The second category includes objects that share partial similarities, such as Gear 1, CouplingHalf, and Gear 2, as well as all assembled objects. As depicted in \cref{fig:dataset} use cases 3, classes such as Gear 1, CouplingHalf, and Gear 2 exhibit a similarity of around 50\%. Therefore, the ConvNext model could potentially misclassify Gear 1 as either CouplingHalf or Gear 2. Additionally, the degree of similarity among assembled objects may be even higher, depending on the size difference between the objects being assembled together. For example, as demonstrated in use cases 5 and 6 in \cref{fig:dataset}, the screw and Orings only constitute a small portion of the Wheel and Power Take Off, respectively. This could potentially bias the model because larger objects may have more easily identifiable features and significant impacts on the model than smaller objects. This could explain the poor results obtained from the confusion matrix in \cref{fig:Confusion_matrix} (e) and (d),  where the model struggled to capture the alignment relationship between the assembled parts and misclassified them into a single category.

The low-performance categories in our dataset may indicate the limitations and challenges associated with Sim-to-Real industrial parts classification. They offer us opportunities to specialize in addressing the most challenging use cases. Furthermore, the presence of high-performance and low-performance categories suggests that our dataset includes use cases in various levels of difficulty and complexity, making it a potential benchmark for evaluating future Sim-to-Real classification models. 

\section{Limitation and Future Work}
\label{sec:futurework}

Given that Sim-to-Real industrial parts classification presents challenges for categories sharing the same albedo and those that are partially the same, our next step is to further develop the dataset to address these challenges.

One possible approach is to randomize the albedo on synthetic images. Adding more variations of color, texture, and material to the CAD models may generate synthetic images with more variations, enabling the network to perceive real-world albedo as just another variation. Moreover, to improve the classification of assembled parts, we could apply different random albedos to each object that is assembled with others. This could potentially allow the models to identify each assembled object and learn their alignment relationships.

In addition, our SIP-17 dataset, which comprises only 17 objects from six industrial use cases, would only partly explain a comprehensive representation of the diverse range of real-world industrial parts. To address this limitation, we intend to increase the size of our dataset by including more isolated and assembled parts with varying degrees of similarity, based on the insights gained from this study.

\section{Conclusion}
\label{sec:conclusion}

In this study, we present a Synthetic Industrial Parts dataset (SIP-17) designed for Sim-to-Real industrial parts classification. It contains 17 objects from six industrial use cases, comprising both isolated and assembled parts. We generated synthetic images using domain randomization techniques, resulting in two datasets: Syn\_R, with randomized backgrounds and postprocessing, and Syn\_O, without them.

To benchmark the dataset, we evaluated it with various state-of-the-art classification models. The models allowed varying levels of performance when training on data from different use cases, with some achieving more than 90\% accuracy while some below 50\%. These results may reveal some potential and challenges of using synthetic data for industrial parts classification and for further creating larger-scale synthetic datasets. One of the main challenges was raised from the subcategories that share similar albedo or are partially the same.

We wish to encourage researchers to focus on Sim-to-Real classification using only synthetic data for training, with a particular emphasis on addressing the challenges posed by the subcategories. Such research has the potential to bring significant benefits to the manufacturing industry, where parts from the same stations often share similar albedo and shapes. Enabling training without real-world data can in principle eliminate the need for data collection and annotation, thus saving time and resources for manufacturers. We hope our work will serve as a preliminary testbed and benchmark for future Sim-to-Real industrial parts classification research.
\section*{Acknowledgement}
\label{sec:Acknowledgement}

This work is partially supported by the Wallenberg AI, Autonomous Systems and Software Program (WASP) funded by the Knut and Alice Wallenberg Foundation. The computations were enabled by resources provided by the Swedish National Infrastructure for Computing (SNIC), partially funded by the Swedish Research Council through grant agreement no. 2018-05973, as well as by the Berzelius resource provided by the Knut and Alice Wallenberg Foundation at
the National Supercomputer Centre.

We gratefully acknowledge our colleagues at the Production Oskarshamn, Production Zwolle, Transmission Assembly, Engine Assembly, Academy, and Smart Factory Lab Departments at Scania CV AB for providing the CAD models and use cases. We extend our thanks to Pooja Rangarajan and her colleagues at the Volkswagen Virtual Engineering Lab for their support in generating synthetic images from CAD models. In addition, we express our sincere appreciation to Marcos Escudero-Viñolo at Universidad Autónoma de Madrid for his insightful suggestions on benchmarking the dataset. 

{\small
\bibliographystyle{ieee_fullname}
\bibliography{reference}
}

\newpage
\section{Supplementary Appendix}
\label{sec:appendix}

Please refer to \cref{TableA1} and \cref{TableA2} for the numerical results corresponding to \cref{fig:all}

\begin{sidewaystable*}
  \centering
  \resizebox{\linewidth}{!}{%
  \begin{tabular}{*{27}{c}}
    \toprule
     &&&&Use case 1 &&&&& &Use case 2& &&&&&Use case 3 &&&& & Use case 4\\
    \midrule
     &&Airgun&Electricity&Hammer&Hook&Plug&\noindent\textbf{Total}& &Fork1&Fork2&Fork3&\noindent\textbf{Total}& &CouplingHalf&Gear1&Gear2&Pinion&\noindent\textbf{Total}& &Cross&Pin1&Pin2&\noindent\textbf{Total}\\
 &ResNet&89.7&72.7&85.3&25.0&98.1&76.2& &78.1&93.3&90.0&87.0&
 &81.8&32.4&86.8&81.8&71.8& 
 &90.0&79.5&83.3&84.3\\
  &EfficientNet&84.6&88.6&91.2&30.0&79.3&74.8&
&68.8&96.7&100.0&88.0&
&60.6&61.8&79.0&43.2&60.1&
&70.0&56.4&83.3&69.6\\
  
  Syn\_R&ConvNext&89.7&95.5&91.2&37.5&98.1&\textcolor{purple}{83.3}&
  &87.5&93.3&100.0&\textcolor{purple}{93.5}&
  &84.9&55.9&92.1&77.3&\textcolor{purple}{77.9}&
  &100.0&87.2&83.3&\textcolor{blue}{90.4}\\
  
   &VIT&94.9&54.6&97.1&52.5&88.7&77.1&
   &100&70&96.7&\textcolor{blue}{89.1}&
   &93.9&64.7&94.7&59.1&\textcolor{blue}{77.2}&
    &90.0&89.7&41.7&74.8\\
   
    &DINO&100.0&47.7&91.2&55.5&100.0&\textcolor{blue}{79.0}&
    &96.9&56.7&100.0&84.8&
    &90.9&70.6&68.4&63.6&72.5&
    &100.0&87.5&63.9&\textcolor{purple}{84.4}\\
    \midrule
    &ResNet&71.8&81.8&97.1&25.0&49.1&63.6&&65.6&46.7&96.7&69.6&&45.5&94.1&0.0&0.0&31.5&&60.0&61.5&30.6&51.3\\
    &EfficientNet&41.0&84.1&64.7&65.0&7.6&50.0&&75.0&56.7&50.0&60.9&&27.3&47.1&71.1&2.3&35.6&&62.5&64.1&50.0&59.1\\
    Syn\_O&ConvNext&97.4&97.7&64.7&65.0&83.0&82.4&&93.8&80.0&100.0&91.3&&97.0&17.7&92.1&59.1&66.4&&92.5&64.1&88.9&81.7\\
    &VIT&97.4&70.5&55.9&37.5&71.7&67.1&&96.9&53.3&93.3&81.5&&100.0&47.1&50.0&56.8&62.4&&92.5&41.0&72.2&68.7\\
    &DINO&100.0&90.9&29.4&52.5&66.0&69.0&&81.3&76.7&100.0&85.9&&93.9&14.7&92.1&6.8&49.7&&100.0&30.8&50.0&60.9\\
    
    \bottomrule
  \end{tabular}
  }
  \caption{Class-wise accuracy (\%) of use cases 1 to 4.Purple and blue colors indicate the best and second-best models in terms of total accuracy trained with Syn\_R.}
  \label{TableA1}
\end{sidewaystable*}

\begin{sidewaystable*}
  \centering
  \resizebox{\linewidth}{!}{%
  \begin{tabular}{*{12}{c}}
    \toprule
 &&&&Use case 5&&&&&Use case 6 \\
    \midrule
    &&Back\_wheel\_with\_screw&Back\_wheel\_without\_screw&Front\_wheel\_with\_screw&Front\_wheel\_without\_screw&\noindent\textbf{Total}& &Orings\_on&TopOring\_off&no\_TopOring&\noindent\textbf{Total}\\
    &ResNet&40.6&15.6&6.3&96.9&39.8&&42.9&54.8&28.6&\textcolor{purple}{42.1}\\
    &EfficientNet&28.1&78.1&3.1&90.6&50.0&&14.3&57.1&47.6&39.7\\
    Syn\_R&ConvNext&78.1&50.0&25.0&59.4&\textcolor{blue}{53.1}&&95.2&21.4&2.4&39.7\\
    &VIT&25.0&100.0&15.6&34.4&43.8&&7.1&31.0&83.3&\textcolor{blue}{40.5}\\
    &DINO&31.3&96.9&0.0&93.8&\textcolor{purple}{55.5}&&0.0&19.1&97.6&38.9\\
    \midrule
    &ResNet&0.0&100.0&0.0&0.0&25.0&&28.6&73.8&0.0&34.1\\
    &EfficientNet&3.1&87.5&25.0&37.5&38.3&&16.7&64.3&38.1&39.7\\
    Syn\_O&ConvNext&46.9&59.4&15.6&71.9&48.4&&97.6&7.1&4.8&36.5\\
    &VIT&0.0&31.3&0.0&96.9&32.0&&4.8&7.1&81.0&31.0\\
    &DINO&0.0&90.6&0.0&90.6&45.3&&2.4&16.7&92.9&37.3\\

    \bottomrule
  \end{tabular}
  }
  \caption{Class-wise accuracy (\%) of use cases 5 and 6.Purple and blue colors indicate the best and second-best models in terms of total accuracy trained with Syn\_R.}
  \label{TableA2}
\end{sidewaystable*}

\end{document}